\documentclass[shortAfour,sageapa,times]{sagej2}

\usepackage{moreverb,url}

\usepackage[colorlinks,bookmarksopen,bookmarksnumbered,citecolor=red,urlcolor=red]{hyperref}

\usepackage[inline]{enumitem}
\usepackage{diagbox}
\usepackage[acronym]{glossaries}
\newacronym{daa}{DAA}{dental age assessment}
\newacronym{fdi}{FDI}{Fédération Dentaire Internationale}
\newacronym{uns}{UNS}{Universal Numbering System}

\newacronym{iofos}{IOFOS}{International Organization for Forensic Odonto-Stoma\-to\-logy}
\newacronym{abfo}{ABFO}{American Board of Forensic Odontology}

\newacronym{cnn}{CNN}{Convolutional Neural Network}
\newacronym{llms}{LLMs}{Large Language Models}

\newacronym{opg}{OPG}{orthopantomography}
\newacronym{dss}{DSS}{decision support system}
\newacronym{iri}{IRI}{Internationalized Resource Identifier}

\newacronym{owl}{OWL}{Web Ontology Language}
\newacronym{rdf}{RDF}{Resource Description Framework}
\newacronym{ttl}{TTL}{Terse RDF Triple Language}

\newacronym{obi}{OBI}{Ontology for Biomedical Investigations}
\newacronym{iao}{IAO}{Information Artifact Ontology}
\newacronym{foaf}{FOAF}{Friend of a Friend}
\newacronym{dcmi}{DCMI}{Dublin Core Metadata Terms}
\newacronym{mls}{ML-Schema}{Machine Learning Schema}
\newacronym{snomed}{SNOMED-CT}{Systematized Nomenclature of Medicine - Clinical Terms}
\newacronym{ohd}{OHD}{Oral Health and Disease Ontology}

\newacronym{ai}{AI}{artificial intelligence}
\newacronym{cq}{CQ}{competency question}

\glsdisablehyper

\newcommand\BibTeX{{\rmfamily B\kern-.05em \textsc{i\kern-.025em b}\kern-.08em
T\kern-.1667em\lower.7ex\hbox{E}\kern-.125emX}}

\begin{document}

\runninghead{Marcelo et al.}

\title{AIdentifyAGE Ontology for Decision Support in Forensic Dental Age Assessment}

\author{%
Renato Marcelo\affilnum{1,2},
Ana Rodrigues\affilnum{3,4},
Cristiana Palmela Pereira\affilnum{3,4},
António Figueiras\affilnum{1,2},
Rui Santos\affilnum{5,4},
José Rui Figueira\affilnum{6,2},
Alexandre P Francisco\affilnum{1,2} and
Cátia Vaz\affilnum{1,7}}

\affiliation{%
\affilnum{1}INESC-ID Lisboa, R. Alves Redol 9, 1000-029, Lisbon, PT\\
\affilnum{2}Instituto Superior Técnico, Universidade de Lisboa, Av. Rovisco Pais 1, 1049-001, Lisbon, PT\\
\affilnum{3}Faculdade de Medicina Dentária da Universidade de Lisboa, Rua Professora Teresa Ambrósio, 1600-277, Lisbon, PT\\
\affilnum{4}FCiências.ID - Centro de Estatística e Aplicações da Universidade de Lisboa (CEAUL), Campo Grande, edifício C1, 3.º, 1749-016, Lisbon, PT\\
\affilnum{5}Instituto Politécnico de Leiria - Escola Superior de Tecnologia e Gestão, R. do Alto Vieiro, Apt. 4163 Edifício D, 2411-901, Leiria, PT\\
\affilnum{6}IST-ID, Av. de António José de Almeida 12, 1000-043, Lisbon, PT\\
\affilnum{7}Instituto Superior de Engenharia de Lisboa, Instituto Politécnico de Lisboa, Rua Conselheiro Emídio Navarro 1, 1959-007, Lisbon, PT}

\corrauth{Cátia Vaz, Instituto Superior de Engenharia de Lisboa, Rua Conselheiro Emídio Navarro 1, 1959-007, Lisbon, PT}

\email{cvaz@cc.isel.ipl.pt}

\begin{abstract}
Age assessment is crucial in forensic and judicial decision-making, particularly in cases involving undocumented individuals and unaccompanied minors, where legal thresholds determine access to protection, healthcare, and judicial procedures.
Dental age assessment is widely recognized as one of the most reliable biological approaches for adolescents and young adults, but current practices are challenged by methodological heterogeneity, fragmented data representation, and limited interoperability between clinical, forensic, and legal information systems.
These limitations hinder transparency and reproducibility, amplified by the increasing adoption of AI-based methods.
The AIdentifyAGE ontology is domain-specific and provides a standardized, semantically coherent framework, encompassing both manual and AI-assisted forensic dental age assessment workflows, and enabling traceable linkage between observations, methods, reference data, and reported outcomes.
It models the complete medico-legal workflow, integrating judicial context, individual-level information, forensic examination data, dental developmental assessment methods, radiographic imaging, statistical reference studies, and AI-based estimation methods.
It is being developed together with domain experts, and it builds on upper and established biomedical, dental, and machine learning ontologies, ensuring interoperability, extensibility, and compliance with FAIR principles.
The AIdentifyAGE ontology is a fundamental step to enhance consistency, transparency, and explainability, establishing a robust foundation for ontology-driven decision support systems in medico-legal and judicial contexts.
\end{abstract}


\maketitle

\section{Introduction}
In recent years, Europe has experienced significant migratory dynamics, with a marked increase in undocumented migration and asylum seekers, including a substantial number of unaccompanied minors~\shortcite{Pereira2025DentalAgeEstimationMigrationAsylum,pradella2017age}. Portugal, as part of this broader European context, has faced growing legal and forensic challenges related to the estimation of age in individuals lacking valid identification documents~\shortcite{Augusto2021DentalAgeAssessment,Pereira2021CutOffLegalAges}. In such cases, age assessment plays a important role, as it directly influences access to legal protection, social services, healthcare, and the applicable judicial framework. Consequently, age estimation must be conducted in a manner that is scientifically robust, ethically sound, transparent, and legally defensible~\shortcite{Pereira2025BoneDentalAgeCriminalResponsibility,schmeling2016forensic}.

Age assessment procedures are inherently interdisciplinary, involving medical, dental, radiological, legal, and ethical dimensions. From a forensic perspective, \acrfull{daa} is widely recognized as one of the most reliable biological approaches, particularly in adolescents and young adults, due to the well-documented chronology of tooth development~\shortcite{Pereira2026DentalAgeAssessmentChildrenAdolescents,yadava2011dental,Pereira2023DentalAgeAssessmentScoring,pinchi2012comparison}
The tooth therefore constitutes a goal biological marker in forensic age estimation, supporting both manual methods based on developmental staging and nowadays approaches relying on \acrfull{ai} applied to radiographic dental imaging~\shortcite{Pereira2025DentalAgeEstimationMigrationAsylum,murray2024applying,Nushi2025PretrainedVGG16}.

Despite the growing methodological advances in \acrshort{daa}, the practical implementation of these procedures remains challenged by the heterogeneity and lack of harmonization resulting from the coexistence of diverse recommendations, guidelines, and protocols issued by multiple international institutions~\shortcite{Pereira2024DentalAgeScoring}.
Clinical data, radiographic findings, methodological parameters, reference studies, statistical outputs, and legal requirements are therefore documented in heterogeneous formats, using inconsistent terminology and variable semantic structures. This lack of standardization limits interoperability, reduces transparency, complicates expert communication, and imposes significant constraints on the development of reliable \acrfull{dss} for forensic and judicial contexts.

In parallel, the evolution of judicial and medico-legal \acrshort{dss} has highlighted the critical need for formally defined, semantically coherent, and interoperable representations of knowledge. Ontologies have emerged as a central instrument in medical informatics to address these challenges, enabling structured knowledge representation, standardized terminology, explicit relationships between entities, and integration across heterogeneous data sources. Several ontologies developed in adjacent domains—such as biomedical investigations, medical imaging, anatomical structures, and machine learning workflows—provide relevant conceptual foundations that can be leveraged to support forensic applications. However, these ontologies are not specifically designed to capture the full complexity of forensic \acrshort{daa}, particularly at the intersection of biological markers, methodological diversity, \acrshort{ai} outputs, and legal decision-making~\shortcite{bernabe2023use}.

To date, no ontology has been developed with a dedicated focus on forensic \acrshort{daa} that comprehensively integrates judicial context, expert workflows, dental development methods, radiographic data, and \acrshort{ai}–based estimation models within a unified semantic framework. This gap limits the reproducibility, comparability, and transparency of age estimation practices and constrains their integration into \acrshort{dss} intended to assist forensic experts and judicial authorities.

The AIdentifyAGE ontology aims to address this unmet need. It provides a standardized, ontology-based modeling framework specifically tailored to forensic \acrshort{daa}, integrating knowledge representation principles with decision support requirements. By formally defining concepts, relationships, and processes related to \acrshort{daa}—using the tooth as a central biological marker—the ontology aims to enhance consistency, interoperability, and transparency, while supporting the development of robust \acrshort{dss} capable of informing forensic and judicial decision-making in complex medico-legal contexts.


\section{Methods}
The AIdentifyAGE ontology started in the scope of the national research project AIdentifyAGE – “Artificial Intelligence and its Application by Forensic Science Service Providers: Migrant Unidentified Age Estimation” (\href{https://doi.org/10.54499/2024.07444.IACDC}{2024.07444.IACDC}). The project aims to support forensic specialists in performing accurate, transparent, and reproducible \acrshort{daa} in undocumented individuals, migration and asylum seekers, addressing both medico-legal and ethical requirements.

The ontology constitutes a core component of the  semantic and knowledge representation framework of the project and provides the formal foundation required for ontology‑driven decision‑support mechanisms in forensic \acrshort{daa}. Its primary objective is to provide a formally defined, semantically coherent, and interoperable model of the data, processes, and outcomes involved in \acrshort{daa}, enabling consistent interpretation, integration, and reuse across systems.

The ontology integrates heterogeneous and multi-source data domains commonly involved in forensic \acrshort{daa}. These include individual\-level information (e.g., reported age, biological sex, and case identifiers), legal and forensic examination data (e.g., requesting authority, examination date, and forensic expert role), and imaging data, with particular emphasis on \acrfull{opg}, the most widely used radiographic modality in \acrshort{daa}.
All entities, terms and relationships, namely their hierarchy, object properties and data properties, as well as their classification and annotation, were obtained in collaboration with domain experts and from de analysis of 10,000 \acrshort{opg} and related medico-legal \acrshort{daa} exams.
By formally representing these domains and their relationships, the ontology supports structured data integration and provides a robust foundation for computational reasoning and decision support.

\begin{figure}[!t]
    \centering
    \includegraphics[width=0.8\linewidth]{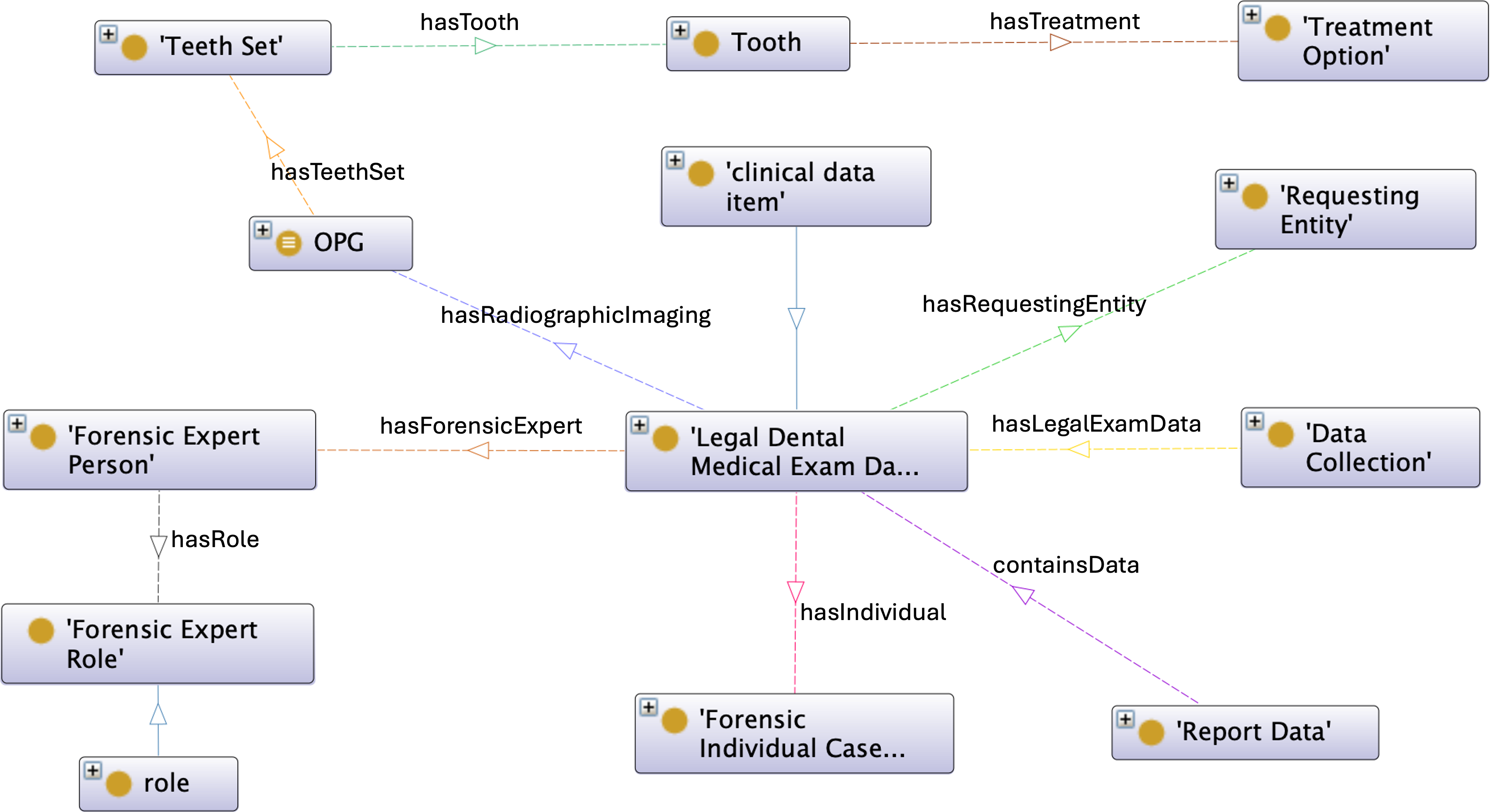}
    \caption{Main entities of AIdentifyAGE judicial/forensic domain. This includes information related to a legal dental medical exam (\textit{Legal Dental Medical Exam Data}) performed by a forensic expert on an undocumented individual. At the end of all legal procedures, also include information regarding the judicial report (\textit{Report Data}), containing the age assessment conclusion.  The non labeled arrows define rdfs:subClassOf property relations. The labeled ones correspond to specific object properties.}
    \label{fig:onto-base}
\end{figure}

\begin{figure}[!t]
    \centering
    \includegraphics[width=\linewidth]{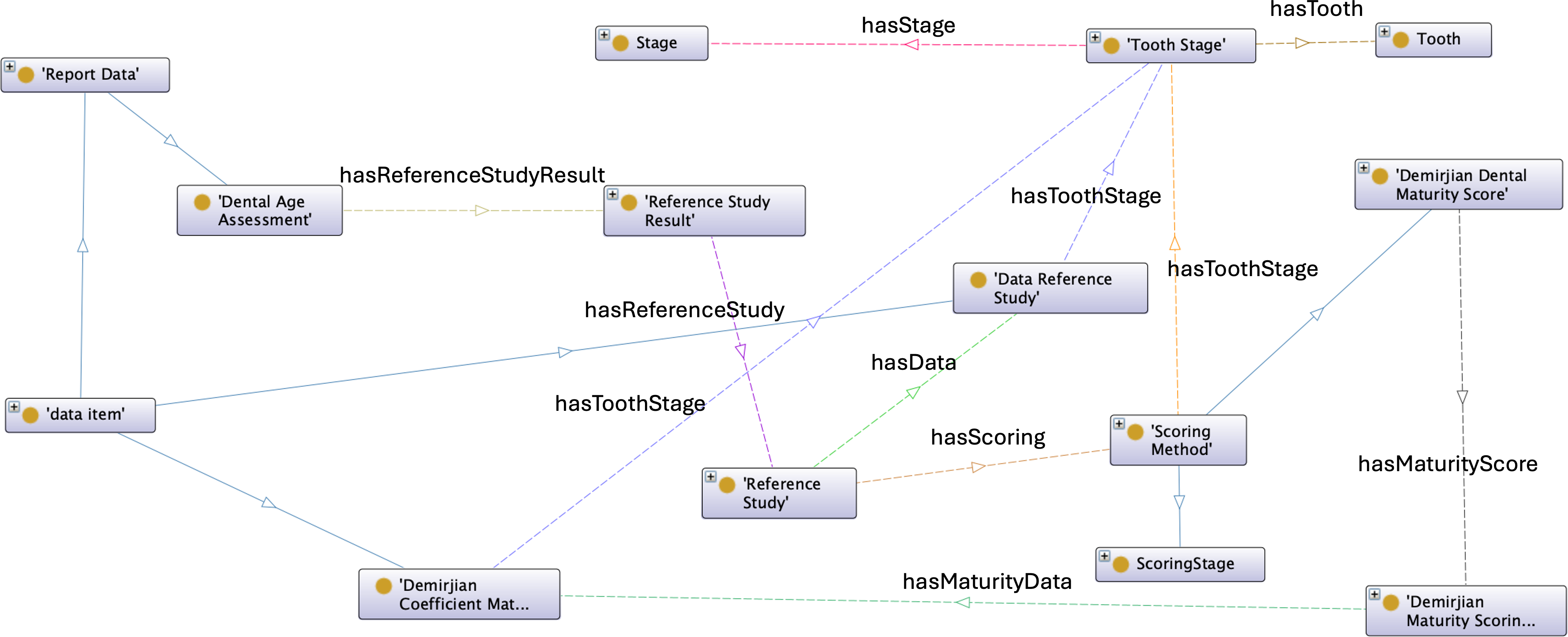}
    \caption{Main entities of AIdentifyAGE manual \acrshort{daa} domain. This includes information regarding \textit{Tooth} development stage scoring (\textit{Tooth Stage}). Given that a set of \textit{Tooth} is scored, a set of \textit{Reference Study} is applied to calculate statistical measures to produce a \textit{Dental Age Assessment} conclusion. \textit{Data Reference Study} and \textit{Coefficient Maturity Data} contain the statistically significant information that allows the \acrshort{daa} to be performed.
    The non labeled arrows define rdfs:subClassOf property relations. The labeled ones correspond to specific object properties.
    }
    \label{fig:onto-manual-scoring}
\end{figure}

\begin{figure}[t]
    \centering
    \includegraphics[width=\linewidth]{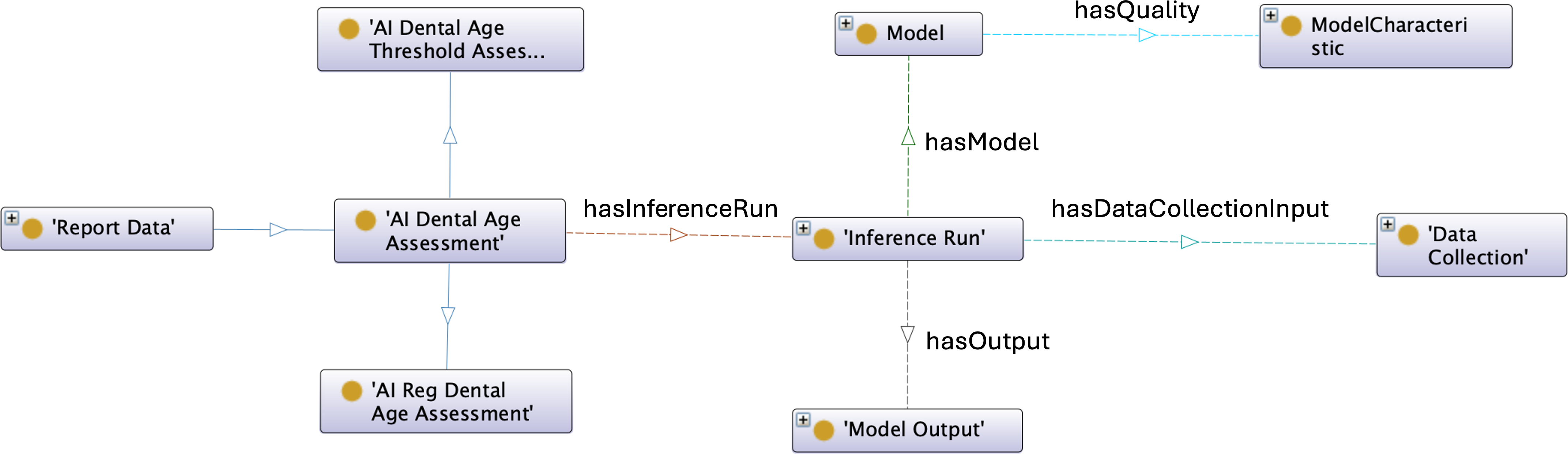}
    \caption{Main entities of AIdentifyAGE \acrshort{ai} \acrshort{daa} domain. This includes information regarding the use of machine-learning models (\textit{Model}) to perform two types of \acrshort{daa}: classification (\textit{\acrshort{ai} Dental Age Threshold Assessment}) and regression (\textit{\acrshort{ai} Reg Dental Age Assessment}). These models were configured using the hyper-parameterizations included in \textit{ModelCharacteristic}, to perform inference (\textit{Inference Run}) over a set of images present in \textit{Data Collection}, producing multiple \textit{Model Output}. The non labeled arrows define rdfs:subClassOf property relations. The labeled ones correspond to specific object properties.}
    \label{fig:onto-ai-scoring}
\end{figure}

The AIdentifyAGE ontology is structured into three domains
\begin{enumerate*}[label=(\roman*)]
\item Judicial/ Forensic, modeling the judicial and forensic process (depicted in \figurename~\ref{fig:onto-base});
\item Manual Dental Age Assessment domain, the \acrshort{daa} process  (illustrated in \figurename~\ref{fig:onto-manual-scoring});
\item \acrshort{ai}‑based Dental Age Assessment domain, modeling machine‑learning \acrshort{daa} process  (outlined in \figurename~\ref{fig:onto-ai-scoring}).
\end{enumerate*}

The ontology explicitly models the methodological components of \acrshort{daa}, complementing the case‑level and imaging metadata represented in \figurename~\ref{fig:onto-base}. As shown in \figurename~\ref{fig:onto-manual-scoring}, clinically and judicially accepted \acrshort{daa} methods are captured through a structured representation of tooth developmental stages, following stage‑based approaches routinely applied in medico‑legal practice. These methods correspond to internationally recognized reference frameworks used for children, adolescents, and young adults—potentially up to 23 years of age—and support the association of developmental stages with population‑specific reference data and statistically derived outputs, including age intervals, mean estimated age, minimum age, and standard deviation, as well as their use in classification tasks based on legally relevant age thresholds~\shortcite{cfa1963age,Demirjian1973,liversidge2008timing,lee2009development}.

The ontology also incorporates age estimation approaches based on \acrshort{ai}, represented in \figurename~\ref{fig:onto-ai-scoring}. It models the key elements of \acrshort{ai}‑driven workflows, including data collections, model characteristics, inference processes, and model outputs, enabling the representation of both classification‑based and regression‑based age assessment models applied to dental radiographic images. This structure supports the integration of outputs from contemporary machine‑learning models—such as convolutional neural networks applied to \acrshort{opg}  images—within a unified semantic framework, ensuring traceability and interpretability of results.
The inherent terms and relationships were derived from analysis of several \acrshort{ai}-based \acrshort{daa} methods (see Karcioglu et al.~\shortcite{karcioglu2025advancing} for a comprehensive survey), including some developed and tested in the context of the project mentioned above, focusing also the necessary statistical assessment measures and tests.
By integrating the forensic, clinical, methodological, and computational components depicted across Figures~\ref{fig:onto-base}, \ref{fig:onto-manual-scoring} and~\ref{fig:onto-ai-scoring}, the ontology provides a structured knowledge base that supports decision‑making in forensic \acrshort{daa}. Its formal representation of data, methods, and outcomes enables consistent interpretation of results and establishes the foundation for an ontology‑based \acrshort{dss} capable of assisting forensic experts and judicial authorities in complex medico‑legal contexts.

All terms and consequently the ontology, were defined following the FAIR principles: Findability, Accessibility, Interoperability, and Reusability \shortcite{wilkinson_fair_2016}. Each identified term was attributed a \acrfull{iri}, and metadata information was added to describe its significance. This ontology was also made publicly available using the BioPortal platform, in broadly applicable formats. During its development, the necessity to allow extensions and/or reusability was taken into account.

\begin{figure}
    \centering
    \includegraphics[width=\linewidth]{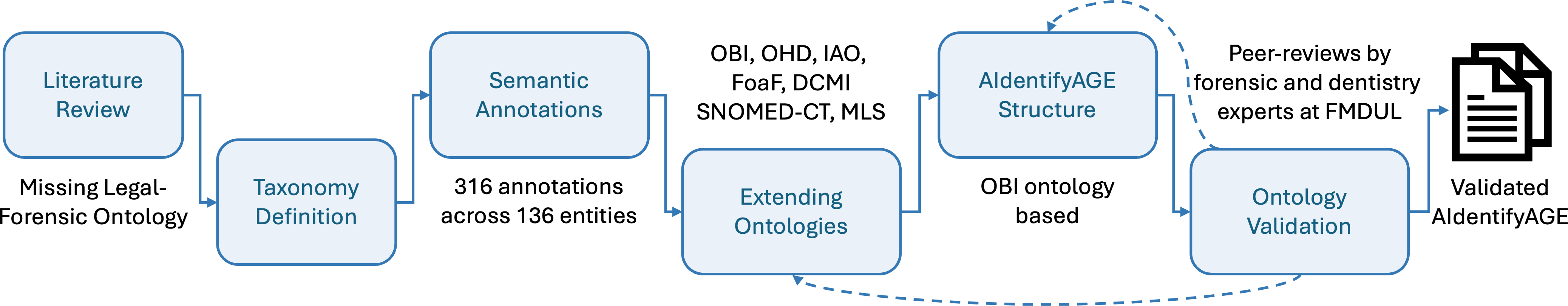}
    \caption{AIdentifyAGE creation process. Starting from literature revision, going through the taxonomy creation, annotation, and hierarchical classification following the OBI ontology, ending with the ontology (iterative) validation, producing the validated ontology.}
    \label{fig:pipeline}
\end{figure}

The AIdentifyAGE ontology is designed to model the complete forensic and legal workflow associated with age assessment in undocumented individuals. Its development proceeded through three structured phases, namely
\begin{enumerate*}[label=(\roman*)]
\item ontology knowledge-base creation,
\item ontology definition chain, and
\item ontology validation process, as illustrated in   \figurename~\ref{fig:pipeline}.
\end{enumerate*}

\subsection{Ontology knowledge-base creation}\label{subsec:onto-kb-create}
This step established the scope of the ontology and defined the characteristics of the terms to be included. The initial categorization relied on the widely adopted class structure of \acrfull{obi}~\shortcite{obi}, which served as the upper‑level framework. Complementary ontologies were subsequently incorporated to represent domain‑specific concepts not covered by OBI, thereby ensuring a coherent and interoperable knowledge base.

\subsection{Ontology definition chain}\label{subsec:onto-def-chain}
After defining the complete knowledge base, the following steps were performed to detail and enrich the current knowledge base:
\begin{enumerate*}[label=(\roman*)]
\item insertion of semantic and scientific meaning to the identified terms,
\item identification of similarities between terms, and
\item linking to the appropriate external ontologies.
\end{enumerate*}

\subsubsection{Insertion of semantic and scientific meaning}
This step was carried out by assigning semantic annotations to each term, including formal definitions, usage notes, and domain‑specific descriptions. These annotations were informed by established glossaries in forensic medicine, dentistry, and biomedical informatics (ISO 18374:2025~\shortcite{ISO18374_2025}, ISO/IEC 23053~\shortcite{ISO23053_2022}, \acrfull{iofos} recommendations~\shortcite{IOFOS2018}, \acrfull{abfo} guidelines~\shortcite{ABFO2023}), as well as by expert consensus when standardized definitions were unavailable. This ensured that every term in the ontology was associated with a clear and scientifically grounded meaning.

\subsubsection{Identification of similarities between terms}
During this step, terms were systematically compared to detect semantic overlap and redundancy. When two terms represented the same concept with identical meaning and constraints, they were merged into a single canonical term. When terms differed in scope or structure but were still conceptually related, equivalence relations were defined to preserve their semantic alignment. In cases where two labels referred to the exact same concept, the most current or domain‑appropriate term was selected as the preferred label (e.g., age assessment instead of age estimation).

\subsubsection{External ontologies linking}
As some of the relevant knowledge domains are already defined in other publicly available ontologies, this step involved linking them to our own. Due to the standardization efforts done while developing and making these ontologies available, no standardization process was needed while integrating them. 

\subsection{Ontology validation process}\label{subsec:onto-valid-process}

Beyond structural and expert validation, involving both the development team and domain experts from the Faculty of Dental Medicine of the University of Lisbon (FMDUL), the AIdentifyAGE ontology was evaluated with respect to its suitability for decision-support applications in medico-legal contexts. Logical consistency was verified using the HermiT reasoner~\shortcite{glimm2014hermit}, confirming satisfiability of all classes and absence of contradictory axioms across reused ontologies.

Functional adequacy was assessed through competency questions representing forensic and judicial information needs. Each \acrfull{cq} was mapped to ontology entities and relations, and verified to be answerable via SPARQL queries over the ontology. A total of 11 \acrshort{cq}s were defined (see \href{https://github.com/AIdentifyAGE/ontology}{ontology GitHub page} for \href{https://github.com/AIdentifyAGE/ontology?tab=readme-ov-file#competency-questions}{the complete list} and for \href{https://github.com/AIdentifyAGE/ontology?tab=readme-ov-file#competency-questiondriven-sparql-queries}{the corresponding SPARQL queries}). This ensures that the ontology supports traceable retrieval of examination context, methodological parameters, statistical outputs, and \acrshort{ai} model provenance. We provide also in the Supplementary Materials a systematic traceability matrix linking each \acrshort{cq} to the ontology classes and properties that enable its formal representation and querying. This mapping demonstrates that the ontology design is fully requirement-driven and that each \acrshort{cq} is grounded in explicit axiomatic definitions.

In addition, interoperability was evaluated by verifying correct semantic alignment with external ontologies, ensuring that reused classes preserve their original semantics and can be queried consistently across integrated knowledge bases. This multi-layer validation confirms the ontology’s readiness to serve as a semantic backbone for forensic \acrshort{daa} decision-support systems. The interoperability of AIdentifyAGE with major biomedical and \acrshort{ai} ontologies is demonstrated through \href{https://github.com/AIdentifyAGE/ontology?tab=readme-ov-file#interoperability-mapping-with-external-ontologies}{an explicit concept-level mapping} provided also in the \href{https://github.com/AIdentifyAGE/ontology}{ontology GitHub page}.

\subsection{Availability}\label{subsec:availability}
The AIdentifyAGE ontology is published in the BioPortal repository following open-source principles and it can be accessed and downloaded locally from \url{https://bioportal.bioontology.org/ontologies/AIDENTIFYAGE}.
Here, the user can find the ontology in the \acrfull{owl}~\shortcite{owl2} standard format. Visiting the ontology GitHub page \url{https://github.com/AIdentifyAGE/ontology}, \acrfull{rdf}~\shortcite{rdf11} and \acrfull{ttl}~\shortcite{ttl} formats are also available to download.

\section{Results}


\begin{table}[ht]
    \centering
    \caption{Ontologies metadata information. Each ontology information refers only to the determined classes, object, and data properties (on AIdentifyAGE) and the relevant subset (on the remaining ontologies).}
    \begin{tabular}{lccc}
        \hline
        \diagbox{Ontology}{Entities} & Classes & Object Properties & Data Properties \\
        \hline
         AIdentifyAGE & 73 & 32 & 47 \\
         \acrshort{obi}~\shortcite{obi} & 36 & 28 & 0\\
         \acrshort{iao}~\shortcite{iao} & 11 & 3 & 0\\
         \acrshort{foaf}~\shortcite{foaf} & 3 & 0 & 3 \\
         \acrshort{dcmi}~\shortcite{dcmi} & 3 & 0 & 3 \\
         \acrshort{mls}~\shortcite{mls} & 25 & 13 & 1 \\
         \acrshort{snomed}~\shortcite{snomed} & 30 & 7 & 0 \\
         \acrshort{ohd}~\shortcite{ohd} & 1289 & 28 & 2 \\ \hline
    \end{tabular}
    \label{tab:ex_onto_details}
\end{table}

As mentioned previously, the AIdentifyAGE taxonomy mainly uses \acrshort{obi}~\shortcite{obi} as an upper ontology, even though it reuses other ontologies as it is depicted in Table~\ref{tab:ex_onto_details}. The most relevant entities of \acrshort{obi} that are more relevant to AIdentifyAGE are:
\begin{enumerate*}[label=(\roman*)]
\item \textit{data item}, \item \textit{clinical data item}, \item \textit{role}, \item \textit{plan specification}, \item \textit{data set}, \item \textit{material anatomical entity}, \item \textit{datum label}, and \item \textit{measurement datum}. 
\end{enumerate*}
For example, \textit{Legal Dental Medical Exam Data} was added as subclass of \textit{clinical data item} (subclass of \textit{data item}); the subclasses \textit{Data Collection}, \textit{Demirjian Maturity Scoring}, and \textit{Reference Study} were mapped as subclass to  \textit{data set} (subclass of \textit{data item}); the \textit{Scoring Method}, \textit{Stage}, and \textit{Treatment Option} classes were mapped as subclass to \textit{plan specification} class; \textit{Teeth Set} class was included under the \textit{mouth} class from \textit{material anatomical entity}; \textit{Forensic Expert Role} was included into the \textit{role} class; the \textit{Tooth Stage} class was included under the \textit{categorical measurement datum} class, while the \textit{Reference Study Result} was included in its superclass, \textit{measurement datum}; classes like \textit{Report Data}, \textit{Demirjian Coefficient Maturity Data}, and \textit{Data Reference Study}, were included under the \textit{data item} class, as none of its subclasses was adequate.

Other AIdentifyAGE classes were classified under other extending ontologies. For example, \textit{Forensic Expert Person} and \textit{Forensic Individual Case Person} were classified under the \acrshort{foaf} ontology, given that both represent a person; \textit{Model Output} and \textit{Inference Run} were classified following the \acrshort{mls} ontology, under \textit{Information Entity} and \textit{Process} class, respectively; \textit{\acrshort{opg}} was classified following the \acrshort{snomed} ontology, under the \textit{Radiographic imaging procedure (procedure)} class. The AIdentifyAGE ontology relies on the \acrshort{mls} ontology~\shortcite{mls} to include information related to the training, testing, and validation of machine-learning models used for performing \acrshort{daa}.

Given the context, \textit{Legal Dental Medical Exam Data} corresponds to the root class with information related to a given legal medical exam. This includes properties regarding the forensic case identification, requesting entity, relevant radiographic imaging, the forensic expert, and the individual undergoing the legal medical examination. \textit{Forensic Expert Person} corresponds to the forensic expert performing the legal medical exam, while the \textit{Forensic Individual Case Person} corresponds to the person undergoing same exam. From a given orthopantomography (\textit{OPG}), we can obtain a \textit{Teeth Set}, aggregating a set of \textit{Tooth}, that can contain (or not) an associated \textit{Treatment Option}.

Regarding manual \acrshort{daa}, each \textit{Tooth} is annotated with a tooth developmental stage according to a selected reference method (e.g., Demirjian~\shortcite{Demirjian1973}, Haavikko~\shortcite{Haavikko1970}, Kullmann~\shortcite{Kullman1992}, or Moorrees–Fanning–Hunt~\shortcite{Moorrees1963}), denoted as \textit{ToothStage} instances. A \textit{Scoring Method} is modeled as the method-specific rule set used to assign stages (and, where applicable, alpha-numeric scores) to teeth and to aggregate tooth-level information at the individual level. The ontology supports multiple scoring schemes by representing:
\begin{enumerate*}[label=(\roman*)]
    \item the staging system (set of permissible stages and their definitions),
    \item optional stage-to-score mappings when a method uses alpha-numeric values, and 
    \item the aggregation procedure used to derive method outputs from tooth-level inputs.
\end{enumerate*}

The statistical interpretation of the assessed stages/scores is modeled through \textit{Reference Study} entities that capture population-specific reference data and their associated parameters. Application of a given \textit{Reference Study} to an assessed case yields a \textit{Reference Study Result}, which stores the derived outputs required for medico-legal reporting and decision support (e.g., age interval, minimum/maximum, mean estimated age, standard deviation), and, when relevant, supports classification relative to legally defined age thresholds.

Regarding \acrshort{ai}-assisted \acrshort{daa}, \textit{Inference Run} is the root class, corresponding to a run performed by a \textit{Model} (e.g., \acrshort{cnn} model) over one or more \acrshort{opg}, contained in a \textit{Data Collection}, producing a \textit{Model Output}. A given \textit{Model} is also characterized by its configurations, included in the \textit{ModelCharacteristic} class.

Finally, the \textit{Report Data} class includes variables regarding the \acrshort{daa} report, such as age range, mean age, and standard deviation.
The \textit{Dental Age Assessment} class, the \textit{\acrshort{ai} Threshold Dental Age assessment} class, and the \textit{\acrshort{ai} Reg Dental Age Assessment} class are used for the manual approach, for the \acrshort{ai}-assisted approach, and for regression tasks, respectively.

It is also important to mention that the \acrshort{ohd} ontology~\shortcite{ohd} already implements the \acrfull{uns}~\shortcite{uns} for each tooth. Given that the numbering system is not standard in all countries, other standard naming schemes were added: \acrfull{fdi} World Dental Federation notation~\shortcite{fdi_notation}, Palmer notation~\shortcite{palmer_notation}, and Haderup notation~\shortcite{haderup-notation}.

\subsection{Term annotations} 
To give more detail and descriptive knowledge to the AIdentifyAGE ontology and its entities, 316 semantic annotations were applied to 70 term classes, 28 object properties, and 58 data properties. Some annotations were added to the ontology to specify information such as versioning and contributors.

These semantic annotations, as well as ontology development, were done using Protégé software~\shortcite{Musen2015Protege}. Each entity received mainly two annotations: a label (to specify the name that should be presented for the entity) and a description (clearly describing the significance of that entity). In cases where a formal definition could not be found, a common agreement between the forensic experts allowed them to reach a non-standard definition.

\subsection{Illustrative use case}
To demonstrate the practical applicability of AIdentifyAGE in a medico-legal context, we present an illustrative forensic \acrshort{daa} use case. The SPARQL query illustrated in Figure~\ref{sparqlListing} retrieves both manual \acrshort{daa} results and \acrshort{ai} model provenance for the illustrative forensic case, including statistical estimates and task type (classification or regression).
\begin{figure}[!t]
  {\small
\begin{verbatim}
PREFIX aida: <https://aidentifyage.github.io/ontology/AIdentifyAGE#>
PREFIX dc:   <http://purl.org/dc/terms/>
PREFIX mls:  <http://www.w3.org/ns/mls#>
PREFIX rdfs: <http://www.w3.org/2000/01/rdf-schema#>
PREFIX xsd:  <http://www.w3.org/2001/XMLSchema#>

SELECT ?meanAge ?stdDev ?minAge ?maxAge ?modelName ?taskType
WHERE {
    # Manual assessment outcome
    ?assessment a aida:DentalAgeAssessment ;
        aida:standardDev ?stdDev ;
        aida:meanAge ?meanAge ;
        aida:minimumProbableAge ?minAge ;
        aida:maximumProbableAge ?maxAge .

    # AI-based assessment provenance
    ?aiAssessment a/rdfs:subClassOf* aida:AIDentalAgeAssessment ;
        aida:hasInferenceRun ?inferenceRun .
    ?inferenceRun a aida:InferenceRun ;
             aida:hasModel ?model .
    ?model a mls:Model .

    ?model mls:hasQuality ?qTask .
    ?qTask dc:title ?title ;
           mls:hasValue ?taskType .
    FILTER(STR(?title) = "task")

    ?model mls:hasQuality ?qName .
    ?qName dc:title ?name ;
           mls:hasValue ?modelName .
    FILTER(STR(?name) = "name")
}
\end{verbatim}
  }
\caption{This illustrative SPARQL query demonstrates how the AIdentifyAGE ontology enables simultaneous retrieval of manual \acrshort{daa} results and \acrshort{ai}‑based assessment provenance within a unified semantic framework. It retrieves statistically outputs from a manual \acrshort{daa}, namely mean estimated age, standard deviation, and age interval, together with information about the \acrshort{ai} model task type used in an \acrshort{ai}‑based assessment 
(classification or regression).}
\label{sparqlListing}
\end{figure}
This example illustrates how AIdentifyAGE enables transparent retrieval of medico-legal conclusions together with methodological and computational context.

\section{Discussion}
AIdentifyAGE addresses a critical gap in forensic medical informatics by providing a formal, interoperable representation of \acrshort{daa} workflows within a legal context. Unlike existing biomedical ontologies that focus on clinical or anatomical aspects in isolation, AIdentifyAGE explicitly models the chain linking observations, methods, reference data, and judicial conclusions.
The integration of \acrshort{ai}-based age assessment within the same semantic framework as manual methods is particularly relevant given increasing regulatory and ethical scrutiny of algorithmic decision-making. By capturing model characteristics, inference processes, and outputs, the ontology supports transparency and explanation, which are key requirements in medico-legal environments.

\subsection{Limitations}
However, ontology-based approaches also introduce limitations. Knowledge acquisition and maintenance require sustained expert involvement, and no ontology can fully eliminate uncertainty inherent to biological age estimation. Moreover, while AIdentifyAGE supports explainability, it does not itself guarantee correctness of underlying models or reference studies; these remain dependent on empirical validation.
Future work should focus on large-scale deployment, integration with institutional information systems, and empirical evaluation of decision-support effectiveness in real forensic workflows.

\subsection{Contributions}
AIdentifyAGE is a reusable, domain-specific ontology model focused on describing the relevant concepts and properties related to forensic and legal procedures supported by \acrshort{daa}, offering a consistent hierarchical concept structure, relevant semantic annotation entities, and accurate relations between terms, serving as the basis for correct data analysis and knowledge handling. It is comprised of 1448 classes, 97 object properties, and 56 data properties, regarding radiographic imaging, odontology, legal and forensic medicine. Its development was closely followed and validated by forensic experts from the Faculty of Dental Medicine - University of Lisbon.

This work makes the following contributions to medical informatics and forensic decision support:
\begin{enumerate*}[label=(\roman*)]
\item We introduce AIdentifyAGE, the first ontology specifically designed to model the complete medico-legal workflow of forensic \acrshort{daa}, integrating judicial context, clinical observations, statistical reference data, and \acrshort{ai}-based inference.

\item We provide a semantically integration of manual and \acrshort{ai}-assisted age assessment methods, enabling explainable linkage between tooth-level observations, population-specific reference studies, and legally relevant age conclusions.

\item We demonstrate how reuse of existent biomedical and machine learning ontologies, aligned under an upper ontology (OBI), enables interoperability and FAIR-compliant knowledge representation in a legal domain.

\item We establish a reusable semantic foundation for ontology-driven decision-support systems intended to assist forensic experts and judicial authorities in age-related determinations.
\end{enumerate*}

\begin{dci}
The authors declare no known potential conflict of interests that could have appeared to influence the work reported in this paper.
\end{dci}

\begin{funding}
  The authors acknowledge the supported by national funds through the Fundação para a Ciência e a Tecnologia, I.P. (FCT) under projects \href{https://doi.org/10.54499/2024.07444.IACDC}{2024.07444.IACDC}, \href{https://doi.org/10.54499/UID/50021/2025}{UID/50021/2025}, \href{https://doi.org/10.54499/UID/PRR/50021/2025}{UID/PRR/50021/2025}, and UID/PRR/00006/2025.
\end{funding}

\bibliographystyle{mslapa}
\bibliography{references.bib}

\end{document}